\documentclass[conference,letterpaper]{IEEEtran}
\IEEEoverridecommandlockouts

\IfFileExists{cite.sty}{\usepackage{cite}}{}
\usepackage{amsmath,amssymb}
\usepackage{graphicx}
\usepackage{booktabs}
\usepackage[hyphens]{url}
\newcommand{\TableText}{\small}

\title{From Complete Scanpaths to Finding-Level \\Gaze Targets in Chest Radiography: \\Structured Cues and Learned Reweighting}

\author{
\IEEEauthorblockN{Sumin Lee$^{1,\dagger}$, Hyunsu Go$^{1,\dagger}$,
Subeen Lee$^1$, Kyeonghun Kim$^2$, and Nam-Joon Kim$^{1,*}$}
\IEEEauthorblockA{
$^1$\textit{Seoul National University}, Seoul, Republic of Korea\\
$^2$\textit{OUTTA}, Republic of Korea\\
\{cirtuare,hsmail02,5584sb,knj01\}@snu.ac.kr;
kyeonghun.kim@outta.ai
}
\thanks{$^{\dagger}$These authors contributed equally to this work.}
\thanks{$^{*}$Corresponding author: Nam-Joon Kim (knj01@snu.ac.kr).}
}

\begin{document}
\IEEEaftertitletext{\vspace{-2\baselineskip}}
\maketitle

\begin{abstract}
Eye tracking records a radiologist's complete scanpath during a reading. A
single reading may contain multiple reported findings, so the scanpath does not
directly provide a separate gaze target for each finding. We construct these
targets by assigning finding-specific weights to the recorded fixations and
rendering them as localization maps. Under a shared rendering protocol, we
compare a structured baseline that combines anatomical, scanpath, temporal,
and directional cues with a learned selector. On 987 REFLACX test instances,
the learned selector's pointing-game accuracy was 0.035 higher than that of the
structured baseline in the primary instance-weighted comparison
(patient-cluster bootstrap 95\% confidence interval, 0.007--0.063). Intersection
over union (IoU) was similar between methods. Structured cues form a strong
baseline, while learned reweighting adds a modest gain in peak
localization.
\end{abstract}

\noindent\textbf{Areas:} Medicine Area 3 (Digital and Precise Medicine);
AI Area 6 (Trustworthy AI).

\section{Introduction}

During chest-radiograph interpretation, eye tracking records a scanpath
synchronized with report dictation~\cite{karargyris2021,reflacx2022}.
Such records have been used to derive spatial supervision for image
models~\cite{stember2021,lanfredi2023}. Because one reading may contain several
findings, however, its scanpath does not directly provide a separate gaze
target for each finding.

Existing approaches use temporal windows to extract label-linked
gaze~\cite{lanfredi2023,ghelichkhan2025}, curate finding-oriented scanpaths, or
predict fixation-level diagnostic intent~\cite{gazesearch2025,radgazeintent2025}.
These studies associate gaze with individual findings, but evaluate downstream
prediction, scanpath generation, or intent classification rather than the
constructed gaze target itself. We retain the complete scanpath, assign
finding-specific weights to its fixations without using image pixels, and
evaluate the rendered map against the corresponding REFLACX region.

\begin{figure*}[!t]
\centering
\includegraphics[width=0.98\textwidth]{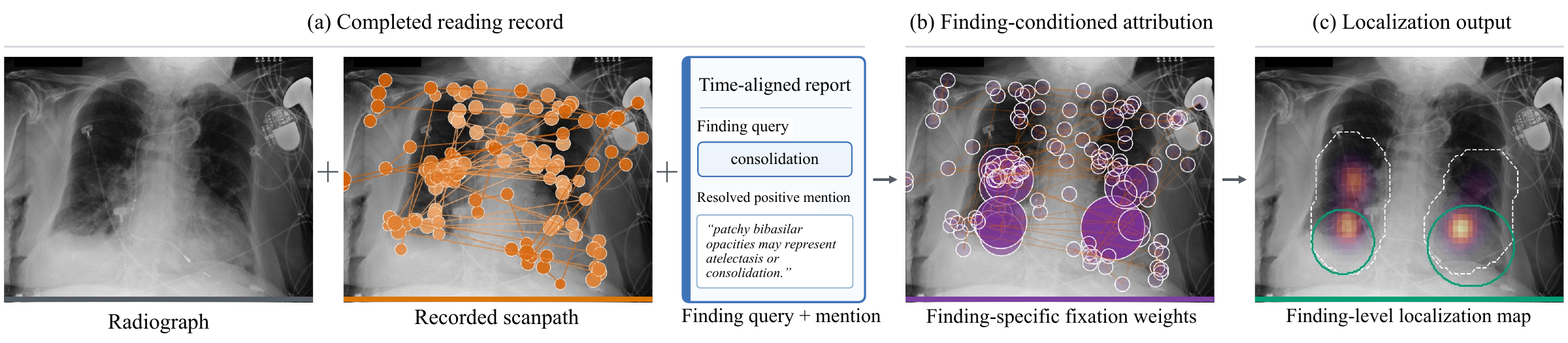}
\caption{Finding-level gaze target construction from a complete REFLACX
reading.
(a) A reported finding and indicators extracted from its linked positive
mentions are paired with the recorded scanpath. (b) A structured or learned
selector assigns finding-specific weights to the fixations. (c) The shared
rendering pipeline
produces a continuous map and a calibrated binary mask (white dashed contour),
which are evaluated against the annotated region (green contour). The
radiograph is shown only for context; image pixels are not inputs to either
selector. Training annotations are used to estimate the anatomical prior and
train the learned selector; test reference regions are reserved for evaluation.}
\label{fig:overview}
\end{figure*}

We ask how well anatomical, scanpath, temporal, and directional cues localize
individual findings and whether learned reweighting improves on this structured
baseline under the same rendering protocol.
Our main contributions are a direct evaluation protocol for finding-level gaze
target construction and a controlled comparison between a structured baseline
and learned reweighting under a shared renderer. We also analyze dependence on
paired gaze--mention records and input features, as well as sensitivity to
training-set size and patient partition.
Figure~\ref{fig:overview} summarizes the task and comparison.

\section{Related Work}

\subsection{Gaze as Supervision in Radiology}

Radiologist gaze has supported image classification and
localization~\cite{saab2021,lanfredi2023,gazegnn2024}, medical image
segmentation~\cite{gazemedseg2024}, and image--text representation
learning~\cite{egma2024}. Recent gaze-aware models include FixationFormer for
chest-radiograph classification~\cite{fixationformer2026}, Gaze-X for reporting,
grounding, and visual question answering~\cite{gazex2026}, and Gaze2Report for
report generation from predicted scanpaths~\cite{gaze2report2026}. These methods
use gaze within downstream prediction or generation; we instead evaluate the
finding-level gaze map constructed from the observed scanpath.

\subsection{Finding-Level Gaze Construction and Evaluation}

Several studies derive gaze representations aligned with report labels or
anatomical regions. Temporal windows produce label-linked maps for downstream
localization~\cite{lanfredi2023,ghelichkhan2025}, while anatomy-centered systems
align report content and gaze to predefined regions~\cite{iai2024,fgcxr2024}.
GazeSearch derives finding-oriented trajectories and evaluates scanpath
prediction~\cite{gazesearch2025}. RadGazeIntent uses images and fixation
sequences to predict diagnostic intent labels~\cite{radgazeintent2025}.
Our protocol retains the complete scanpath and outputs a distribution over its
original fixations. The rendered gaze target is evaluated directly against a
held-out spatial annotation without using image pixels as selector inputs.

Psycholinguistic studies show that gaze can precede verbal
naming~\cite{meyer1998,griffin2000}, motivating pre-mention windows; prior
REFLACX work selected a 1.5-s lookback for label-specific gaze
construction~\cite{lanfredi2023}. Multimodal annotation systems likewise align
spoken tokens with temporally nearby gaze~\cite{stember2021,vaidyanathan2020}.
Temporal alignment alone can include fixations associated with other findings,
whereas an anatomical prior cannot identify which locations were visited in the
target reading. These complementary limitations motivate a structured baseline
that combines mention timing, a training-derived anatomical prior, scanpath
support, and directional terms.

REFLACX ellipses allow finding-level gaze maps to be evaluated independently
of downstream image-model training~\cite{reflacx2022}. Evaluation against these
ellipses measures spatial localization rather than moment-to-moment diagnostic
intent~\cite{hooge2025}.
Pointing-game accuracy assesses peak placement, whereas
intersection over union (IoU) quantifies thresholded spatial
overlap~\cite{zhang2016pointing,saporta2022}. We report both to distinguish
peak localization from map extent.

\section{Methods}
\label{sec:method}

\subsection{Task and Shared Renderer}

An instance consists of a complete reading, a finding label $c$, a fixation
sequence $G=\{g_i=(x_i,y_i,t_i,d_i,v_i)\}_{i=1}^{N}$, a nonempty set
$\mathcal{M}$ of linked positive mentions, and a reference region $E_c$. Here,
$(x_i,y_i)$ are normalized image coordinates, $t_i$ is the fixation midpoint,
$d_i$ is its duration, and $v_i$ is its inter-fixation velocity. The reference
region is the union of the finding's ellipse annotations. A selector outputs
weights $a=(a_1,\ldots,a_N)$ that sum to one over the recorded fixations. The
renderer places these weights at the fixation coordinates,
\begin{equation}
B(z)=\sum_{i=1}^{N}a_i\,K_0(z;(x_i,y_i)),
\label{eq:splat}
\end{equation}
where $K_0$ places a point mass on a $64\times64$ grid. A Gaussian kernel
with a validation-selected bandwidth smooths the resulting map, and a
validation-selected threshold binarizes it for IoU. The pointing-game
prediction is the location of the maximum of the smoothed continuous map.

At test time, selectors receive $c$, $\mathcal{M}$, and $G$; the held-out
reference region is used only for evaluation. The ellipse union serves as a
coarse localization annotation rather than a segmentation mask or a label of
diagnostic intent for each fixation~\cite{reflacx2022}. The training objective
therefore rewards in-region fixations without optimizing the ellipse boundary.

\subsection{Structured Cue Baselines}
\label{sec:structured}

The structured baselines combine four sources of information: an anatomical
prior, directional terms, scanpath support, and mention timing.

\emph{Anatomical prior.} For finding $c$, the training-derived map
\begin{equation}
A_c(z)=\frac{1}{|\mathcal{D}^{\mathrm{tr}}_c|}
\sum_{r\in\mathcal{D}^{\mathrm{tr}}_c}\mathbb{I}[z\in E_{r,c}],
\label{eq:anatomy}
\end{equation}
estimates its typical spatial support. $\mathcal{D}^{\mathrm{tr}}_c$ contains
the mention-linked training instances for finding $c$; validation and test
ellipses do not contribute to this map.

\emph{Directional terms.} For each opposing pair (left/right and upper/lower),
the deterministic modulation $L_{\mathcal{M}}[A_c]$ restricts the prior to the
corresponding image half when one term appears without its opposite.
Half-plane orientation is estimated from training ellipse centers. The prior
is unchanged when neither or both terms in a pair appear and is restored if a
restriction removes all support. Middle, bilateral, and retrocardiac indicators
are used only by the learned selector.

\emph{Scanpath support.} Given a prior map $P$, the selector evaluates it at
the target reading's fixation coordinates and normalizes the resulting scores,
\begin{equation}
w_i(P)=\frac{P(x_i,y_i)}{\sum_j P(x_j,y_j)},
\label{eq:scanpath}
\end{equation}
using uniform weights if every sampled value is zero. Equation~\eqref{eq:splat}
then renders the weighted scanpath.

\emph{Temporal gate.} Let $T_i(\mathcal{M})$ indicate that fixation $i$
overlaps any linked-sentence window. The fixed-window temporal
baseline weights fixation $i$ in proportion to $d_iT_i(\mathcal{M})$. A
temporally gated structured baseline instead uses weights proportional to
$d_iP(x_i,y_i)T_i(\mathcal{M})$. If the gate selects no fixation, the temporal
baselines revert to duration weighting over the full scanpath. If the selected
fixations have no prior support, the structured method uses the temporal
weights alone. The combined structured baseline sets
$P=L_{\mathcal{M}}[A_c]$.

\subsection{Learned Fixation Reweighting}

Each fixation is represented by five temporal and kinematic features: the
signed offset to the nearest linked sentence onset, its absolute value, a
pre-onset indicator, duration, and velocity. When multiple mentions are
linked, the closest onset in absolute time is used. Unlike the fixed-window
baseline, the learned selector receives continuous timing features rather than
a binary window indicator. Offsets and durations are measured in seconds
and velocity in normalized-image units per second; these features are not
standardized.
Normalized fixation coordinates are encoded with sine and cosine functions at
frequencies $\pi\{1,2,4,8\}$ for each axis, yielding 16 positional
features~\cite{tancik2020}.

We extract seven spatial indicators by matching predefined lexical terms in
the linked sentences: left, right, bilateral, upper, lower, middle, and
retrocardiac. Three additional indicators represent low-severity or uncertain
language, high-severity language, and diffuse or patchy extent. Corresponding
terms include \emph{mild} or \emph{possible}, \emph{severe} or
\emph{extensive}, and \emph{diffuse} or \emph{patchy}, respectively. Each
binary indicator is active if any corresponding term occurs in a linked
sentence. An 8-dimensional learned finding embedding and the ten mention
indicators are concatenated and projected to a query $q$; the fixation
features are projected to key vectors $k_i$. Both projections have dimension
16.

The prespecified selector uses all ten indicators. A four-indicator control
retains only the left, right, upper, and lower indicators included in the
deterministic rule, with the fixation features, optimization settings, and
calibration procedure unchanged.

The learned selector uses a single scaled dot-product attention
layer~\cite{vaswani2017},
\begin{equation}
a_i=\frac{\exp(k_i^{\top}q/\sqrt d)}
{\sum_j\exp(k_j^{\top}q/\sqrt d)}, \qquad d=16.
\label{eq:attention}
\end{equation}
This single-layer architecture reweights the observed fixations without a
spatial decoder.

Training maximizes attention mass within the reference region while
discouraging collapse onto a single fixation,
\begin{equation}
\mathcal{L}=-\log\!\left(\sum_i a_i\,\mathbb{I}[(x_i,y_i)\in E_c]+\epsilon\right)
-0.01\,H(a).
\label{eq:loss}
\end{equation}
Here, $H(a)$ is the Shannon entropy of the fixation-weight distribution and
$\epsilon=10^{-6}$ provides numerical stability. Instances without a fixation
inside $E_c$ provide no positive target under this loss and are therefore
excluded from training; they remain in validation and test.

\section{Experimental Setup}
\label{sec:setup}

\subsection{REFLACX Cohort and Mention Linking}

We use REFLACX Phase 3~\cite{reflacx2022}, built on
MIMIC-CXR~\cite{mimiccxr2019}. A rule-based linker matches report sentences
using a predefined finding-specific keyword dictionary and excludes nonpositive
contexts marked by variants of \emph{no}, \emph{without},
\emph{resolved}, \emph{clear of}, \emph{free of}, \emph{negative for},
and \emph{rule out}. The mention-linked cohort comprises annotated finding
instances with at least one linked positive mention. The dictionary does not
cover \emph{Other} or \emph{Support devices}, which are excluded from the
mention-linked cohort. Linker coverage uses all annotated finding instances,
including those without a linked mention, as its denominator.

Both datasets are de-identified and available through PhysioNet under
credentialed access~\cite{PhysioNet-reflacx-xray-localization-1.0.0,
PhysioNet-mimic-cxr-2.0.0}. The REFLACX study reports an exemption from the
University of Utah Institutional Review Board, with no informed consent
required for eye-tracking collection. The MIMIC-CXR study reports approval by
the Beth Israel Deaconess Medical Center Institutional Review Board, with
informed consent waived.

When evaluated against the released manually corrected report labels, the
linker achieves 94.8\% recall for positive annotated pairs (273 of 288); all
273 retrieved pairs are positive in the corrected labels. Exact sentence
assignment and mention onset were not manually adjudicated.

Using NumPy seed 0, we deterministically split patients into 55\% training,
15\% validation, and 30\% test cohorts. The mention-linked cohorts contain
1,895 training, 547 validation, and 987 test instances. Of the 1,093
annotated finding instances in the test split before mention linking, 987
(90.3\%) have a linked positive mention and form the evaluation cohort,
spanning 398 patients. Because a reading may contribute multiple findings, all
grouping and resampling are performed by patient.

All mention-linked test instances have nonempty scanpaths, and 99.5\% have at
least one fixation inside the annotated region. Four training instances have no
in-region fixation and are excluded from the training loss. The median scanpath
contains 114 fixations (interquartile range, 84--147).

\subsection{Comparators, Controls, and Operating Points}

The complete-scanpath density baseline assigns weights proportional to fixation
duration and uses neither finding identity nor mention timing. It measures the
localization available from the reading's overall gaze density. The fixed-window
temporal baseline uses a window that begins 1.5 s before linked-sentence onset,
following Lanfredi et al.~\cite{lanfredi2023}, and extends through sentence end,
as in subsequent sentence-level gaze construction~\cite{ghelichkhan2025}. The
anatomical prior and structured cue baselines are defined in
Section~\ref{sec:structured}. The anatomical-prior baseline uses $A_c$ directly,
with its operating point selected on validation data. We use the prespecified
1.5-s window to analyze cue composition; the primary comparison and
training-size analysis use a lookback selected on validation data.

For a lookback $L$, each linked-sentence window begins $L$ seconds before
sentence onset, is clipped at the preceding sentence onset if necessary, and
ends at the linked sentence end. Overlapping intervals are merged when multiple
sentences match. We construct windows for
$L\in\{0.5,1.0,1.5,2.0,3.0\}$ s from the complete record and calibrate each
candidate independently. Validation IoU selects the 3.0-s lookback, the largest
value tested, for the primary structured baseline.

A matched other-patient control replaces each eligible target gaze--mention
record with complete records from training patients that share the finding and
all seven spatial indicators. Substitute records are scored with the target
finding and all ten indicators from the target mentions, while their fixation
timing is defined relative to their own linked mentions. Eligible records
within each exact-match group are ranked deterministically. Validation selects
the number of matched maps to average from
$K\in\{1,2,4,8\}$; all five seeds select $K=8$. The control is evaluated only for
test instances with at least one exact match.

Feature controls use all 987 mention-linked test instances. A
feature-restricted selector is trained with only the finding embedding and
temporal/kinematic features, omitting positional features and mention
indicators. For the ten-indicator selector, one evaluation-time perturbation
permutes the Fourier positional features across fixations, a second permutes
the five temporal/kinematic features, and a third masks the seven spatial
indicators. Both permutations are performed within each record, and the
resulting weights are rendered at the original fixation coordinates.

Training-set sensitivity uses five nested patient-level subsample chains at
10\%, 25\%, and 50\% of the 735 eligible training patients, crossed with five
optimizer seeds. At each fraction, the learned selector and structured
baseline are rebuilt from the same retained patients; the test cohort remains
fixed. We first average seeds within each chain and then across chains. All
conditions retain the full validation cohort for calibration, and each
structured baseline selects its lookback by validation IoU at that training
fraction.

\subsection{Evaluation, Statistical Analysis, and Implementation}

For each map family, we select a Gaussian smoothing bandwidth and binarization
threshold by maximizing validation IoU and fix both values for test evaluation.
Pointing-game accuracy is evaluated with the same bandwidth, without
metric-specific retuning. Evaluation-time feature controls reuse the
ten-indicator selector's calibration; the other-patient control is calibrated
on validation instances with an exact match. Candidate bandwidth and threshold
grids are fixed before validation.

All methods use the same ellipse rasterization and test metrics; fixation-based
methods also share the rendering and upsampling procedure. Training,
validation, and test sets are patient-disjoint. Anatomical priors are estimated
from training ellipses, calibration and architecture selection are performed on
the validation cohort, and final metrics are computed on the test cohort. Results
for the learned selector are averaged over five optimizer seeds on the primary
split. The partition analysis repeats the full five-seed pipeline for each of
five patient partitions.

Pointing-game accuracy records whether the map maximum falls inside
$E_c$~\cite{zhang2016pointing,saporta2022}; ties use the first maximum in
row-major order. IoU measures overlap between the thresholded map and $E_c$ and
therefore depends on smoothing and threshold calibration.

Each reading--finding pair is one evaluation case. For each metric, the primary
estimand is the mean difference across 987 mention-linked cases, calculated as
the learned selector value minus the validation-selected structured baseline
value. The patient-cluster bootstrap resamples patients and retains all
instances from each sampled patient; the 2.5th and 97.5th percentiles of 10,000
resamples define the 95\% confidence intervals
(CIs)~\cite{field2007cluster}. Five-seed
means are computed per instance before resampling. A Wilcoxon signed-rank
sensitivity analysis uses patient-level mean differences, giving each patient
equal weight. Optimizer seeds are repeated computations rather than independent
observations; variation across seeds and patient partitions is summarized
separately.

The learned selector is trained for 40 epochs with a batch size of one using
Adam with a learning rate of $10^{-3}$~\cite{kingma2015adam}.
Both weight initialization and example order are seeded. Maps are rendered at
$64\times64$ and evaluated after nearest-neighbor upsampling to
$256\times256$. Fixations outside the displayed image remain in the sequence
and are clipped to the nearest border cell during rasterization. Using
validation IoU, we select from a $2\times2$ grid of scoring and coordinate
encoding choices: multilayer-perceptron versus cross-attention scoring, and raw
versus Fourier-encoded coordinates. The selected configuration uses
cross-attention with Fourier positional encoding and is fixed before test
evaluation.

\section{Results}

\subsection{Structured Baselines and Primary Comparison}

Table~\ref{tab:structured-baselines} compares single-source baselines,
combinations of structured cues, and learned reweighting. Among the
prespecified 1.5-s constructions, combining the anatomical prior, scanpath
support, directional terms, and mention timing gives the highest pointing-game
accuracy and IoU.

\begin{figure*}[!t]
\centering
\begin{minipage}[t]{0.488\textwidth}
\centering
\scriptsize (a) Pleural abnormality\par
\vspace{2pt}
\begin{minipage}[t]{0.488\linewidth}
\centering
\scriptsize Structured\par
\vspace{2.5pt}
\includegraphics[width=\linewidth]{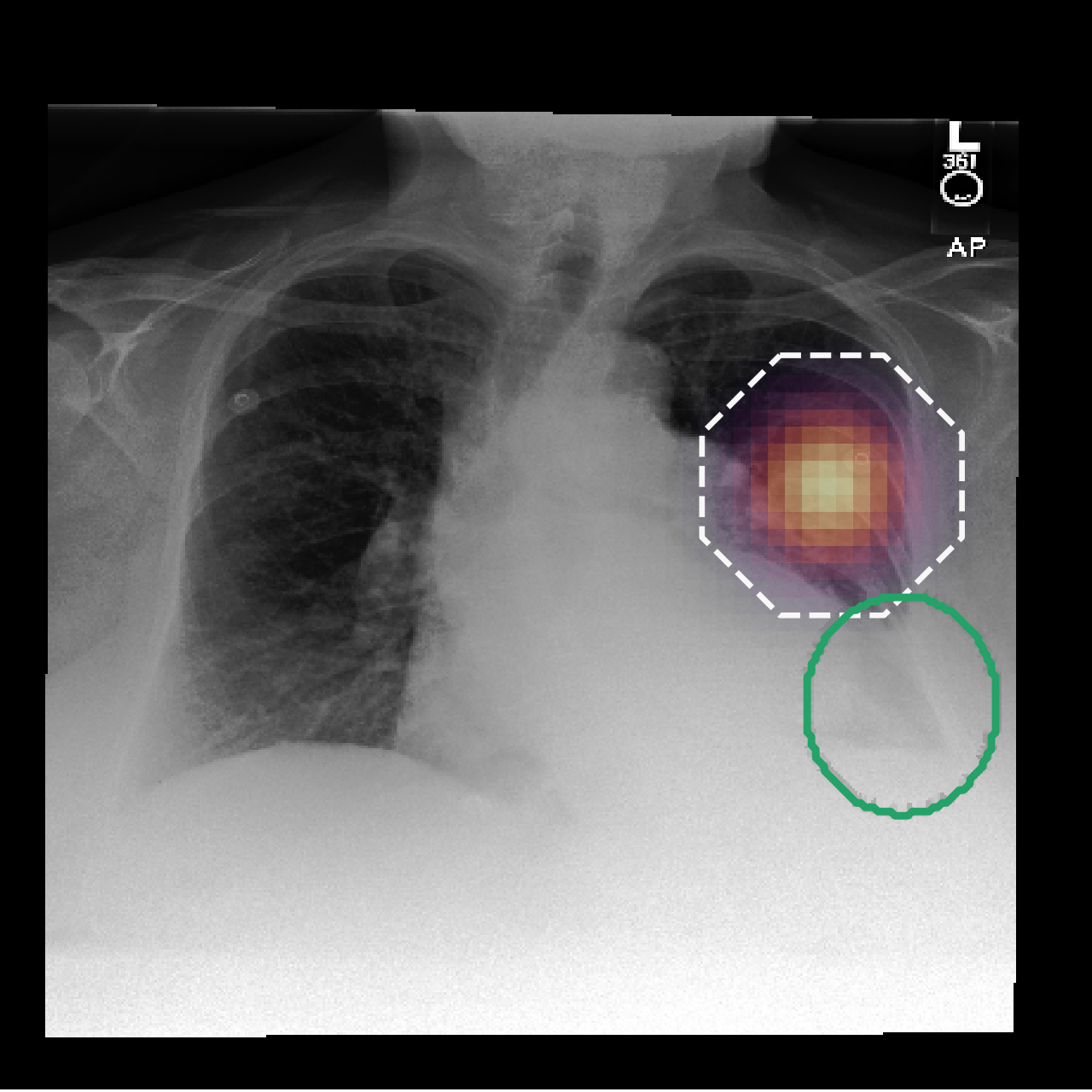}
\end{minipage}\hfill%
\begin{minipage}[t]{0.488\linewidth}
\centering
\scriptsize Learned\par
\vspace{2.5pt}
\includegraphics[width=\linewidth]{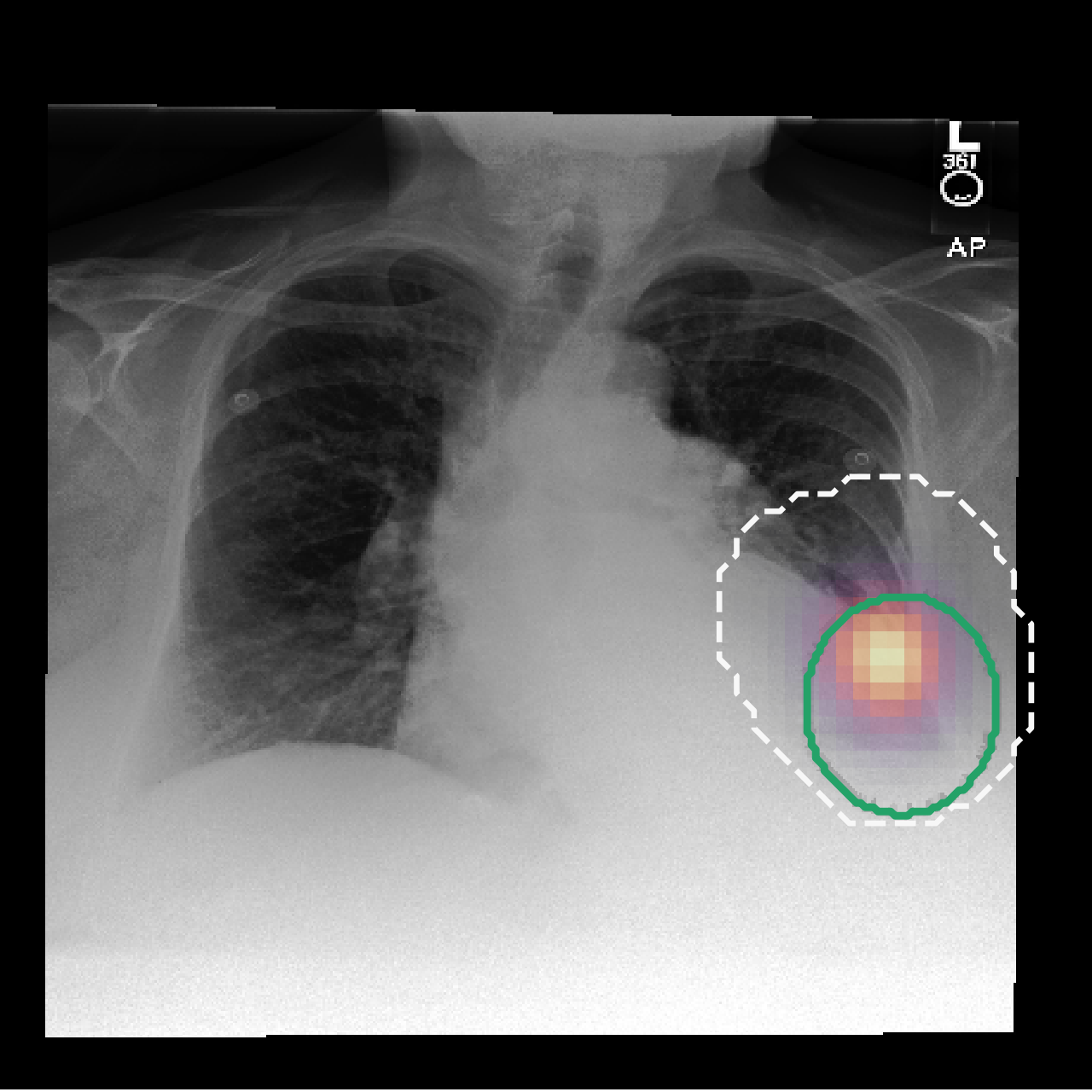}
\end{minipage}
\end{minipage}\hfill%
\begin{minipage}[t]{0.488\textwidth}
\centering
\scriptsize (b) Lung nodule or mass\par
\vspace{2pt}
\begin{minipage}[t]{0.488\linewidth}
\centering
\scriptsize Structured\par
\vspace{2.5pt}
\includegraphics[width=\linewidth]{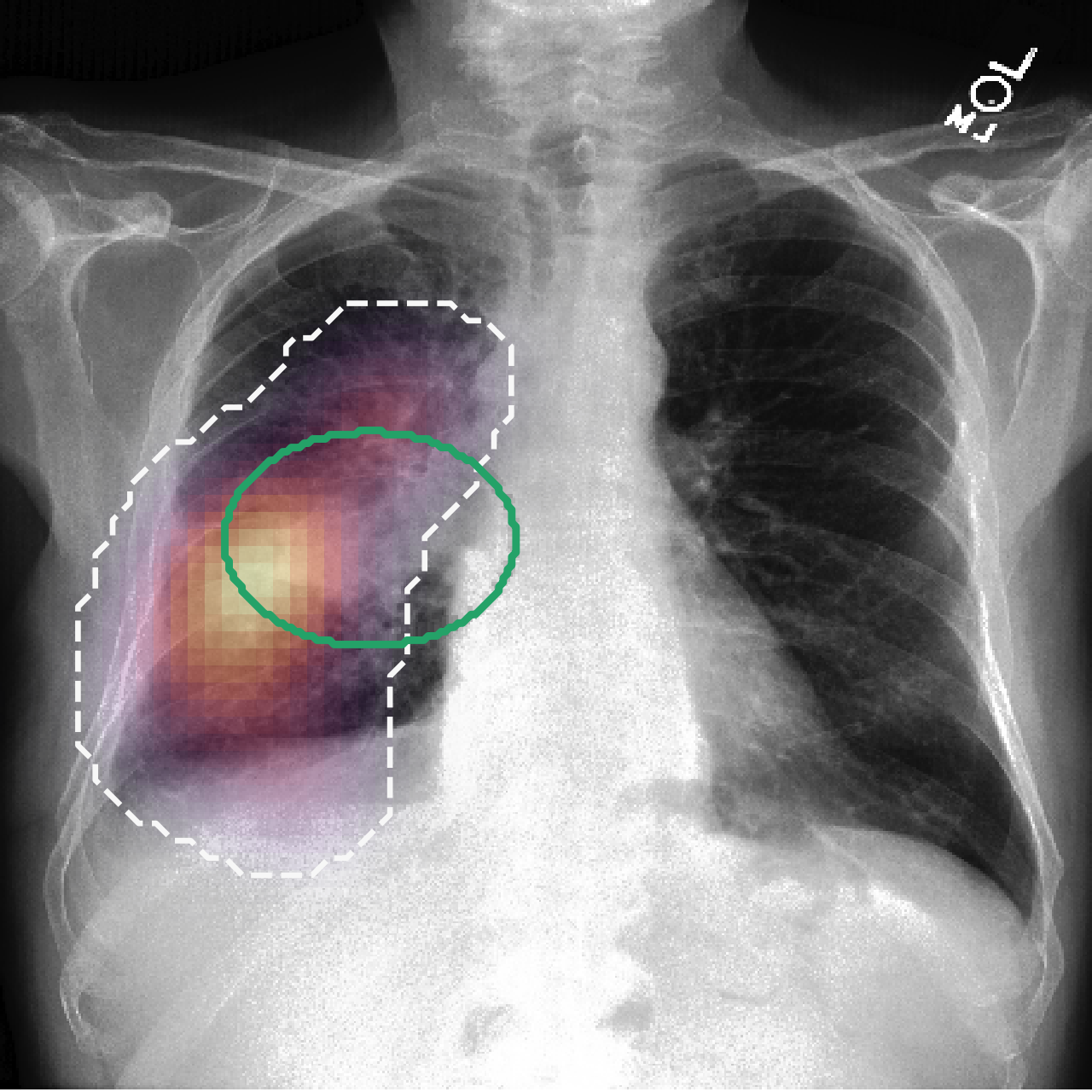}
\end{minipage}\hfill%
\begin{minipage}[t]{0.488\linewidth}
\centering
\scriptsize Learned\par
\vspace{2.5pt}
\includegraphics[width=\linewidth]{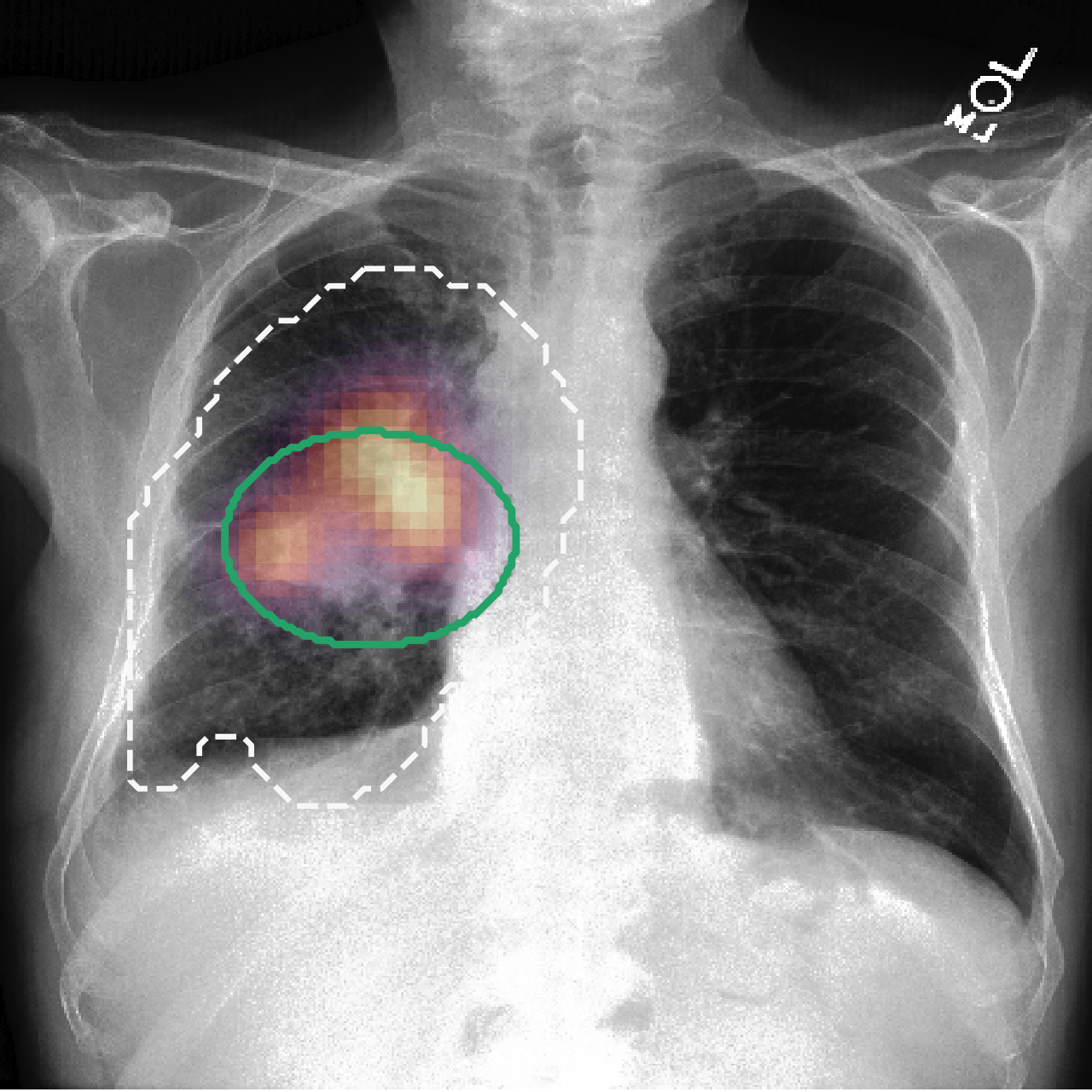}
\end{minipage}
\end{minipage}
\par\vspace{4pt}
\caption{Examples from the primary comparison: (a) pleural abnormality and (b)
lung nodule or mass. In both cases, the validation-selected 3.0-s structured
map peaks outside the reference region, whereas the mean learned map peaks
inside. The structured IoU and five-seed mean learned IoU are 0.006 and 0.386
in (a), and 0.222 and 0.272 in (b). A purple--yellow scale denotes continuous
map intensity, and
green contours mark the reference regions. White dashed contours show the
calibrated structured mask or, for the learned map, the majority-vote mask
across the five seed-specific calibrated masks. Learned maps are averaged
across the five seeds.}
\label{fig:method-comparison}
\end{figure*}

\begin{table}[!b]
\centering
\caption{Localization results on 987 test instances, grouped by construction
strategy. The temporal and 1.5-s structured rows use the prespecified lookback;
the primary comparison uses the 3.0-s structured baseline selected by
validation IoU. Learned values are five-seed means.}
\label{tab:structured-baselines}
\TableText
\setlength{\tabcolsep}{4pt}
\begin{tabular*}{\columnwidth}{@{\extracolsep{\fill}}lcc@{}}
\toprule
Method & Pointing acc. & IoU \\
\midrule
\multicolumn{3}{@{}l}{\emph{Single-source baselines}} \\
Complete-scanpath density & 0.406 & 0.201 \\
Anatomical prior & 0.504 & 0.274 \\
1.5-s temporal baseline & 0.621 & 0.277 \\
\addlinespace
\multicolumn{3}{@{}l}{\emph{Combined structured cues}} \\
Prior $\times$ scanpath support & 0.672 & 0.306 \\
Prior $\times$ scanpath + directional terms & 0.718 & 0.341 \\
Prior $\times$ scanpath + temporal gate & 0.753 & 0.320 \\
1.5-s combined structured baseline & 0.792 & 0.344 \\
3.0-s combined structured baseline & 0.789 & 0.355 \\
\addlinespace
\multicolumn{3}{@{}l}{\emph{Learned reweighting}} \\
Ten-indicator learned selector & 0.825 & 0.358 \\
\bottomrule
\end{tabular*}
\end{table}

\begin{table}[t]
\centering
\caption{Lookback selection and test sensitivity for the combined structured
baseline. Each candidate is calibrated independently; validation IoU selects
3.0 s within the tested grid. Test metrics are reported only for sensitivity
analysis.}
\label{tab:window-sweep}
\TableText
\setlength{\tabcolsep}{2.5pt}
\begin{tabular*}{\columnwidth}{@{\extracolsep{\fill}}lccc@{}}
\toprule
& Validation & \multicolumn{2}{c}{Test} \\
\cmidrule(lr){2-2}\cmidrule(l){3-4}
Lookback (s) & IoU & Pointing acc. & IoU \\
\midrule
0.5 & 0.335 & 0.733 & 0.319 \\
1.0 & 0.353 & 0.776 & 0.335 \\
1.5 & 0.362 & 0.792 & 0.344 \\
2.0 & 0.370 & 0.798 & 0.349 \\
3.0 & \textbf{0.373} & 0.789 & 0.355 \\
\bottomrule
\end{tabular*}
\end{table}

In the test sensitivity analysis, IoU increases across the tested grid, whereas
pointing-game accuracy peaks at 2.0 s (Table~\ref{tab:window-sweep}).

In the primary instance-weighted comparison, the learned selector exceeds the
structured baseline by 0.035 in pointing-game accuracy (patient-cluster
bootstrap 95\% CI, 0.007--0.063) and by 0.004 in IoU (95\% CI,
$-0.003$--0.010).

This instance-weighted analysis indicates improved peak localization, with no
clear difference in thresholded spatial overlap. A separate Wilcoxon
signed-rank sensitivity analysis of patient-level mean
differences yielded $p=0.077$ for pointing-game accuracy and $p=0.842$ for IoU.

Figure~\ref{fig:method-comparison} shows two cases in which the learned map
moves the peak into the reference region.

\subsection{Record and Feature Dependence}

An exact match on the finding and all seven spatial indicators is available for
948 of 987 test instances from 389 patients. On this subset, replacing the
paired record with matched other-patient records reduces pointing-game accuracy
by 0.282 (patient-cluster bootstrap 95\% CI, 0.247--0.316) and IoU by 0.077
(95\% CI, 0.065--0.089). These reductions indicate that matching the finding
and indicators alone does not reproduce the localization obtained from the
paired gaze--mention record. Because the substituted records differ jointly in
fixation coordinates, timing, and underlying lesion geometry, this control does
not isolate any one of these factors.

Table~\ref{tab:feature-controls} probes the selector's inputs on the full
cohort. Restricting the query to the four directional indicators increases
pointing-game accuracy by 0.013 relative to the prespecified ten-indicator
model (patient-cluster bootstrap 95\% CI, 0.001--0.025), while IoU remains
similar (difference $-0.001$; 95\% CI, $-0.004$--0.002). Thus, the six
additional indicators extracted from the mentions do not improve aggregate
performance in this model.
Removing both positional features and mention indicators during training
reduces both metrics, and positional-feature permutation causes the largest
decrease among the evaluation-time perturbations.

\begin{table}[t]
\centering
\caption{Feature controls on 987 test instances. Values are five-seed
means; evaluation-time perturbations reuse the ten-indicator selector's
calibration.}
\label{tab:feature-controls}
\TableText
\setlength{\tabcolsep}{3pt}
\begin{tabular*}{\columnwidth}{@{\extracolsep{\fill}}lcc@{}}
\toprule
Condition & Pointing acc. & IoU \\
\midrule
Ten-indicator selector & 0.825 & 0.358 \\
Four-indicator selector & 0.837 & 0.357 \\
Finding + temporal/kinematic only & 0.730 & 0.314 \\
Positional features permuted & 0.553 & 0.231 \\
Temporal/kinematic features permuted & 0.683 & 0.298 \\
Spatial indicators masked & 0.750 & 0.307 \\
\bottomrule
\end{tabular*}
\end{table}

\subsection{Sensitivity to Training Size and Patient Partition}

Table~\ref{tab:training-size} summarizes training-size sensitivity. The learned
selector trails the structured baseline built from the same patients at 10\%
but exceeds it in pointing-game accuracy from 25\% onward. Its IoU remains lower
at the partial fractions and becomes comparable at 100\%. Because every
condition retains the full annotated validation cohort, these results do not
measure total annotation cost.

\begin{table}[t]
\centering
\caption{Training-size sensitivity across 735 eligible patients. Partial rows
average five seeds within each of five nested subsample chains; the 100\% row
averages five seeds. Structured lookbacks are selected on validation and are
3.0 s at every fraction.}
\label{tab:training-size}
\TableText
\setlength{\tabcolsep}{2.5pt}
\begin{tabular*}{\columnwidth}{@{\extracolsep{\fill}}lcccc@{}}
\toprule
& \multicolumn{2}{c}{Pointing acc.} & \multicolumn{2}{c}{IoU} \\
\cmidrule(lr){2-3}\cmidrule(l){4-5}
Training fraction & Learned & Structured & Learned & Structured \\
\midrule
10\% & 0.768 & 0.778 & 0.319 & 0.346 \\
25\% & 0.803 & 0.790 & 0.335 & 0.352 \\
50\% & 0.819 & 0.789 & 0.342 & 0.353 \\
100\% & 0.825 & 0.789 & 0.358 & 0.355 \\
\bottomrule
\end{tabular*}
\end{table}

Across the five patient partitions, the learned selector exceeds the structured
baseline by 0.017--0.036 in pointing-game accuracy; IoU differences range from
near zero to 0.012.

\section{Discussion and Limitations}

The structured baseline combines typical location with scanpath support,
mention timing, and directional language.

Record substitution and feature controls indicate that the learned selector
relies on case-specific fixation positions; additional mention indicators do
not improve aggregate performance.

Pointing-game accuracy and IoU emphasize different properties. The loss rewards
attention mass on recorded in-region fixations, whereas IoU also depends on
smoothing, thresholding, and the coarse ellipse boundary. Learned reweighting
can therefore move the map maximum without materially changing the thresholded
mask, so peak localization can improve while IoU remains similar.

The evaluated cohort is limited to chest radiographs and findings with a
rule-linked positive mention; negative findings and report-level uncertainty
are not evaluated as separate targets. Linker validation is limited to
report-level positivity; exact sentence assignment and mention onset were not
adjudicated. The reference ellipses are coarse localization annotations rather
than segmentation masks or labels of diagnostic intent for individual
fixations. The released Phase 3 files omit reader identity, precluding
reader-held-out evaluation~\cite{reflacx2022}.

The selected 3.0-s lookback lies at the upper boundary of the tested grid.
Smoothing bandwidths selected by validation IoU are also used for
pointing-game evaluation. The training-size experiment retains a fully
annotated validation cohort and therefore does not establish an advantage in
total annotation cost. Downstream utility of the constructed targets remains
to be tested, and conclusions about learned reweighting are limited to the
single-layer selector evaluated here.

\section{Conclusion}

An anatomical prior, scanpath support, mention timing, and directional terms
form a strong baseline for finding-level gaze target construction from complete
readings. Under the shared rendering protocol, learned reweighting modestly
improves peak localization in the instance-weighted comparison, while IoU
remains similar. Direct evaluation of the constructed maps separates target
construction from downstream image-model training.

\section*{Acknowledgment}

This work was supported by the SNU Student-Directed Education Undergraduate
Research Program through Seoul National University (2026). This work was supported by the IITP AI Semiconductor Program and the ANCHOR Program, funded by the Korean government and the Seoul Metropolitan Government (IITP-2023-RS-2023-00256081; 2026-ANCHOR-01-110).

\section*{Code Availability}

\begingroup
\raggedright
The analysis code is publicly available at
\url{https://github.com/TeamBeaverSudal/cxr-fixation-attribution}.
\par
\endgroup

\bibliographystyle{IEEEtran}
\bibliography{references}

\end{document}